\documentclass[conference]{IEEEtran}
\IEEEoverridecommandlockouts
\usepackage{cite}
\usepackage{amsmath,amssymb,amsfonts}
\usepackage{algorithmic}
\usepackage{graphicx}
\usepackage{textcomp}
\usepackage{xcolor}

\usepackage[linesnumbered,titlenumbered,ruled]{algorithm2e}

\usepackage{booktabs} 
\usepackage{colortbl}
\def\redc{\cellcolor[HTML]{FF999A}}
\def\orangec{\cellcolor[HTML]{FFCC99}}
\def\yellowc{\cellcolor[HTML]{FFF8AD}}

\usepackage{multirow,makecell}

\def\R{\mathbb{R}}

\def\R{\mathbb{R}}

\def\our{CEC-MMR}

\def\BibTeX{{\rm B\kern-.05em{\sc i\kern-.025em b}\kern-.08em
    T\kern-.1667em\lower.7ex\hbox{E}\kern-.125emX}}
\begin{document}

\title{\our{}: Cross-Entropy Clustering Approach to Multi-Modal Regression}

\author{\IEEEauthorblockN{ Krzysztof Byrski}
\IEEEauthorblockA{Jagiellonian University \\
}
\and
\IEEEauthorblockN{Jacek Tabor }
\IEEEauthorblockA{Jagiellonian University}
\and
\IEEEauthorblockN{Przemysław Spurek}
\IEEEauthorblockA{Jagiellonian University}
\and
\IEEEauthorblockN{Marcin Mazur}
\IEEEauthorblockA{Jagiellonian University}
}


\maketitle

\begin{abstract}
In practical applications of regression analysis, it is not uncommon to encounter a multitude of values for each attribute. In such a situation, the univariate distribution, which is typically Gaussian, is suboptimal because the mean may be situated between modes, resulting in a predicted value that differs significantly from the actual data.  Consequently, to address this issue, a mixture distribution with parameters learned by a neural network, known as a Mixture Density Network (MDN), is typically employed. However, this approach has an important inherent limitation, in that it is not feasible to ascertain the precise number of components with a reasonable degree of accuracy. In this paper, we introduce \our{}, a novel approach based on Cross-Entropy Clustering (CEC), which allows for the automatic detection of the number of components in a regression problem. Furthermore, given an attribute and its value, our method is capable of uniquely identifying it with the underlying component. The experimental results demonstrate that \our{} yields superior outcomes compared to classical MDNs.

\end{abstract}


\section{Introduction}

A classical regression method is a statistical technique that is typically utilized to ascertain the relationship between an input (or observation) variable and an output (or response) variable. In contrast, multi-modal regression (also known as multi-output regression) is concerned with the simultaneous prediction of multiple real-valued output variables, which allows for a much broader range of applications. These include (but are not limited to) the modeling of ecosystems \cite{kocev2009using}, chemometric analysis of multivariate calibration \cite{burnham1999latent}, forecasting of the audio spectrum of wind noise \cite{kuznar2009curve}, concurrent estimation of disparate biophysical parameters from remote sensing images \cite{tuia2011multioutput}, and channel estimation from multiple received signals \cite{sanchez2004svm}. In all of the aforementioned applications, multi-output regression methods frequently demonstrate superior predictive performance compared to single-output approaches.


\begin{figure}[ht!]
\begin{tabular}{cc}
\rotatebox[origin=c]{0}{GMM} & \rotatebox[origin=c]{0}{CEC} \\
\makecell{\!\!\!\!\!\includegraphics[width=0.48\columnwidth]{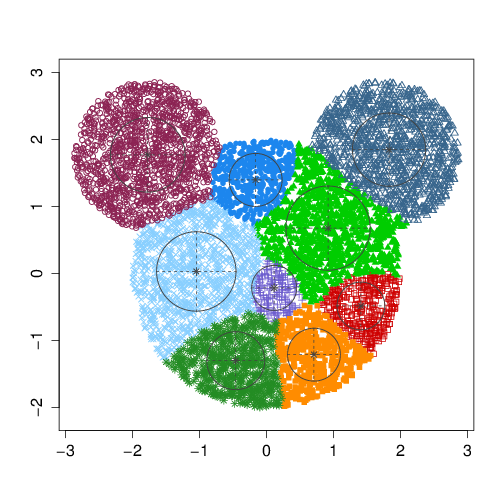}}
& \makecell{\includegraphics[width=0.48\columnwidth]{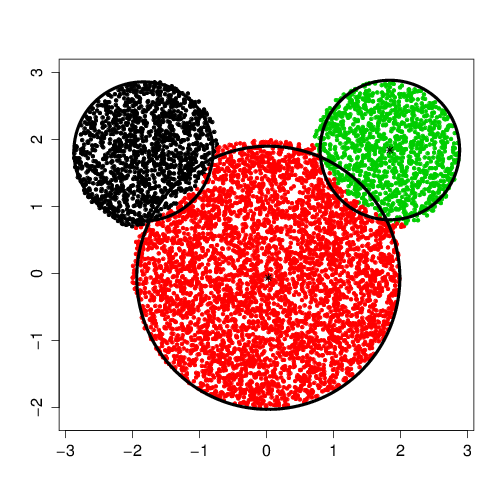}}
\end{tabular}
\caption{Qualitative comparison between Gaussian Mixture Model (GMM) and Cross-Entropy Clustering (CEC) on a toy mouse-like dataset. The results presented were produced with the R packages mclust~\cite{fraley2006mclust} and CEC~\cite{spurek2017r}. The final clustering is illustrated with a variety of colors. It should be noted that in the case of CEC, the initial number of clusters (10) was reduced to 3. The presented example was inspired by \cite{tabor2014cross}.}
\label{img:cec_gmm}
\end{figure}

In multi-modal regression, the description of the output variable based on uni-modal distributions is frequently inadequate, as the mean value may fall between the modes. Therefore, such a predictor is incorrect because it is not feasible to model multiple components simultaneously. As an alternative method, a mixture of distributions with parameters learned by a neural network can be utilized \cite{bishop1994mixture,cui2022gaussian,ellefsen2019mixture,lee1998semiparametric,zen2014deep}. Accordingly, the conditional distribution is modeled by a mixture of density distributions (typically Gaussian). Such an approach, which is known as the Mixture Density Network (MDN), represents an effective means of addressing a range of multi-modal regression problems \cite{ellefsen2019mixture,zen2014deep}. However, this method requires manual specification of the number of components to identify the true data distribution, which presents a significant challenge. This is due to the fact that the number of outputs may fluctuate in accordance with a value of the input variable. In the event that the number of components in the mixture exceeds the number of outputs, the model must attempt to merge the distributions. Conversely, when the number of components is insufficient, it is not possible to accurately describe all potential values within the response variable.

\begin{figure*}[ht!]
	\centering
	\begin{tabular}{@{}c@{}c@{\!\!\!\!}c@{\!\!\!\!}c@{\!\!\!\!}c@{\!\!\!\!}c@{\!\!\!\!}c@{}}
        \rotatebox[origin=c]{90}{\tiny MDN} &
        \makecell{\includegraphics[width=0.18\textwidth]{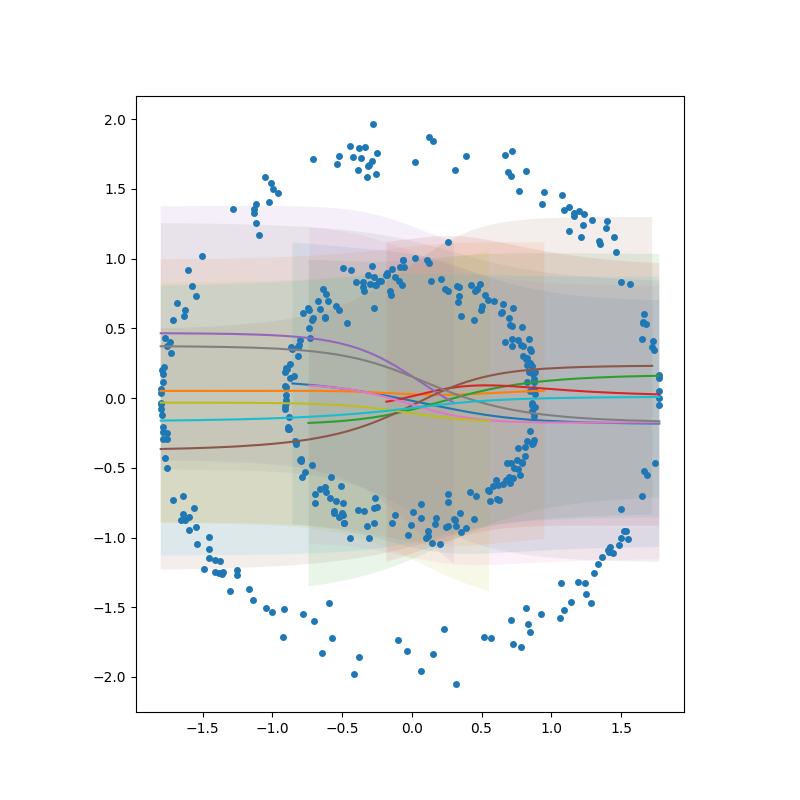}} &
	\makecell{\includegraphics[width=0.18\textwidth]{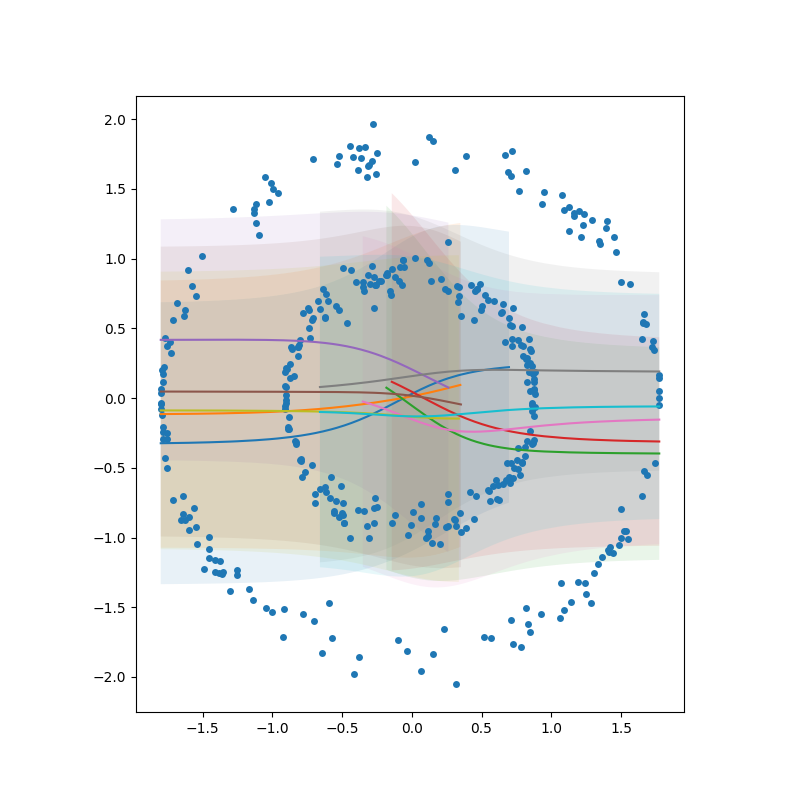}} &
	\makecell{\includegraphics[width=0.18\textwidth]{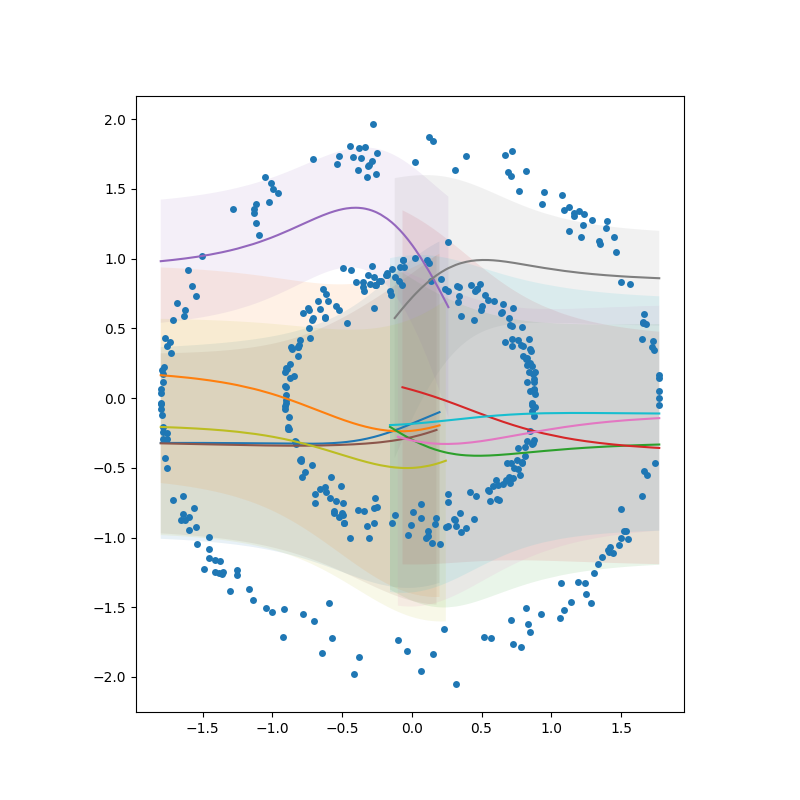}} &
\makecell{\includegraphics[width=0.18\textwidth]{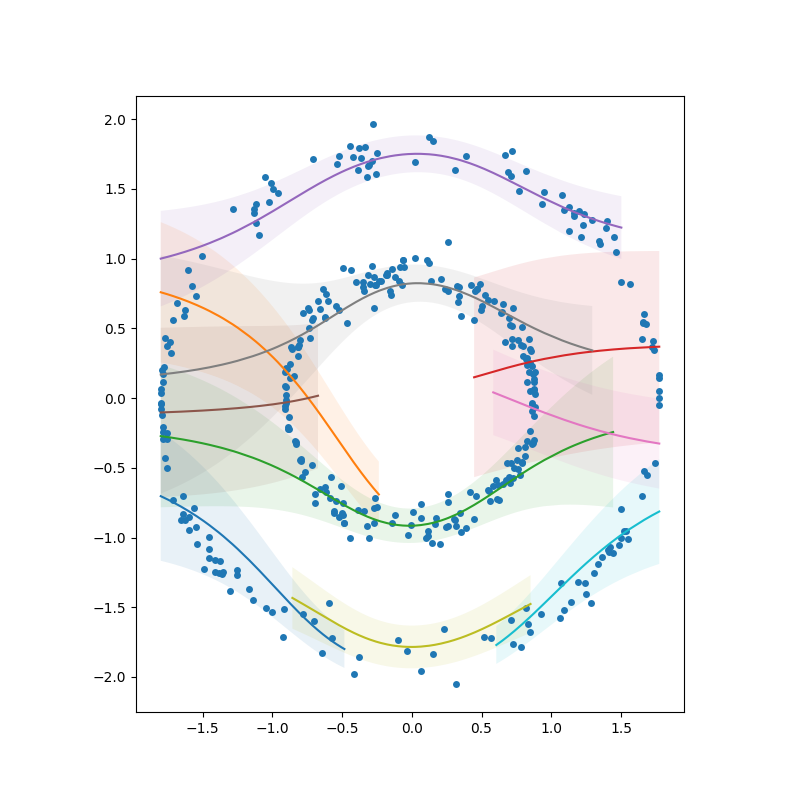}} &
	\makecell{\includegraphics[width=0.18\textwidth]{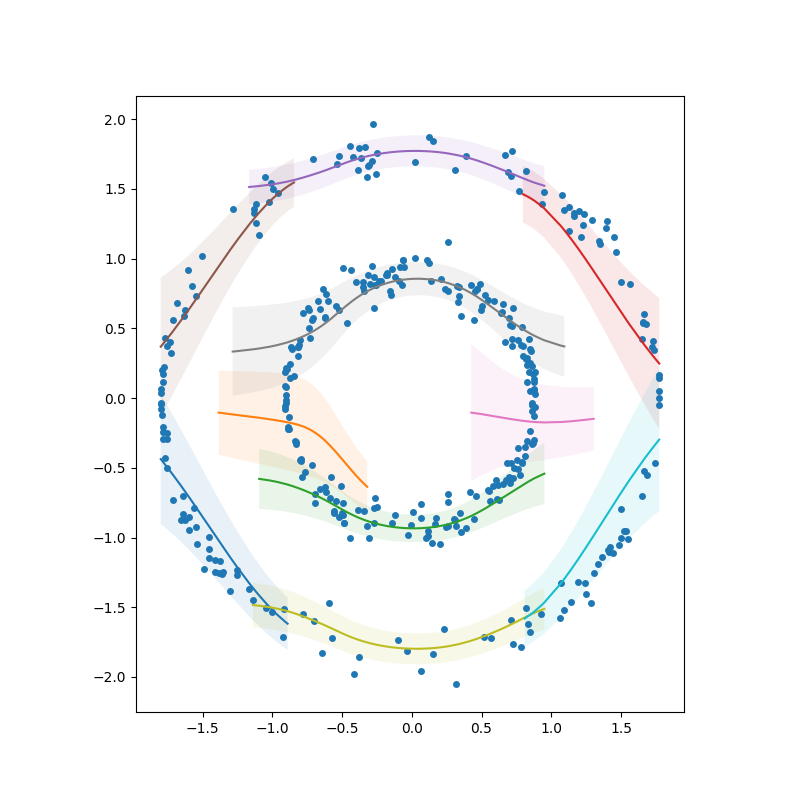}} &
\makecell{\includegraphics[width=0.18\textwidth]{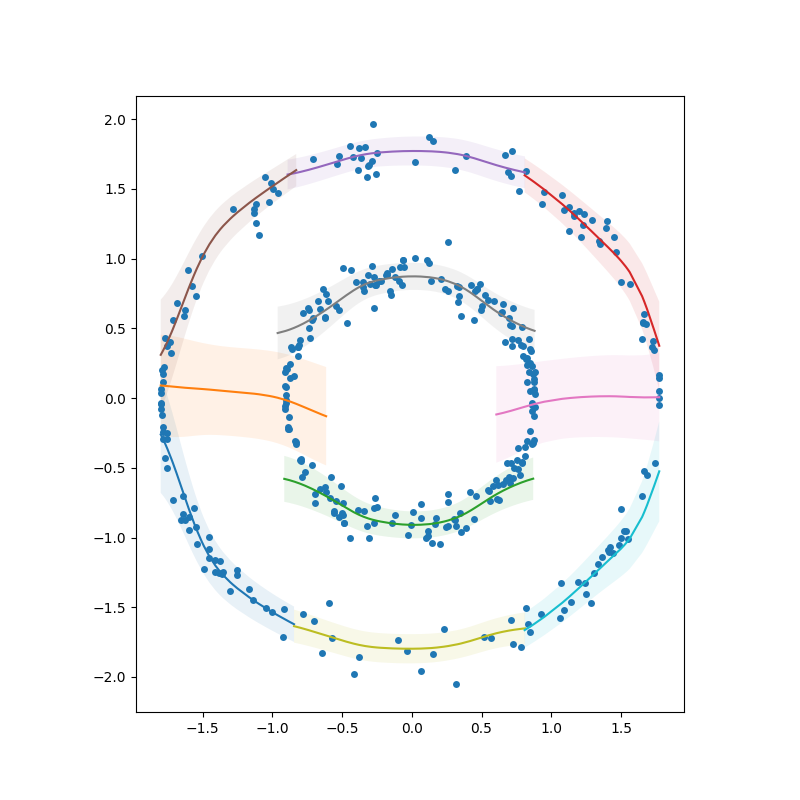}} \\
        \rotatebox[origin=c]{90}{\tiny \ \ \our{} (our)} &
		\makecell{\includegraphics[width=0.18\textwidth]{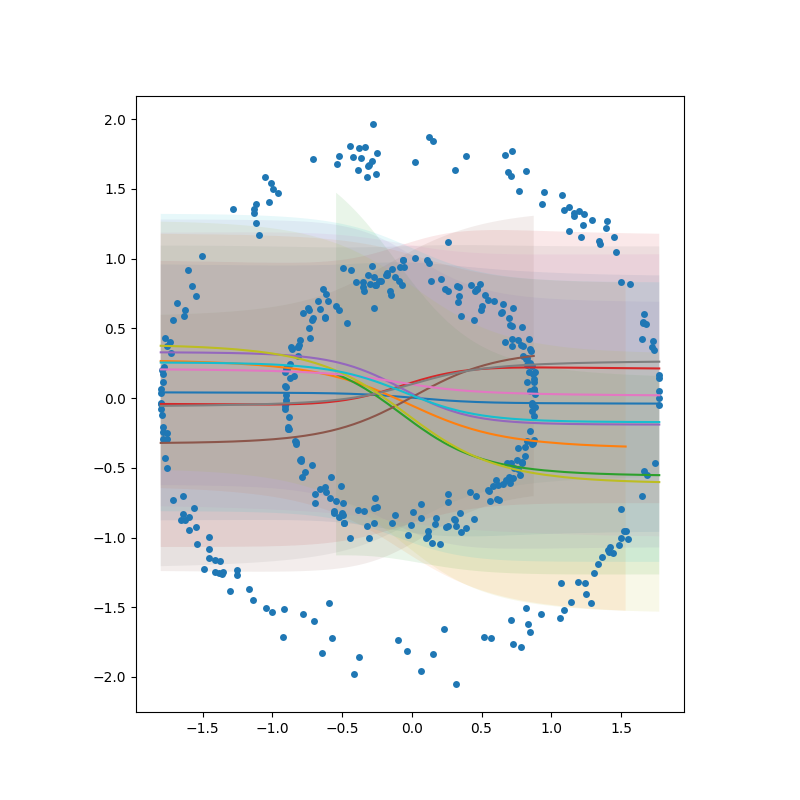}} &
	\makecell{\includegraphics[width=0.18\textwidth]{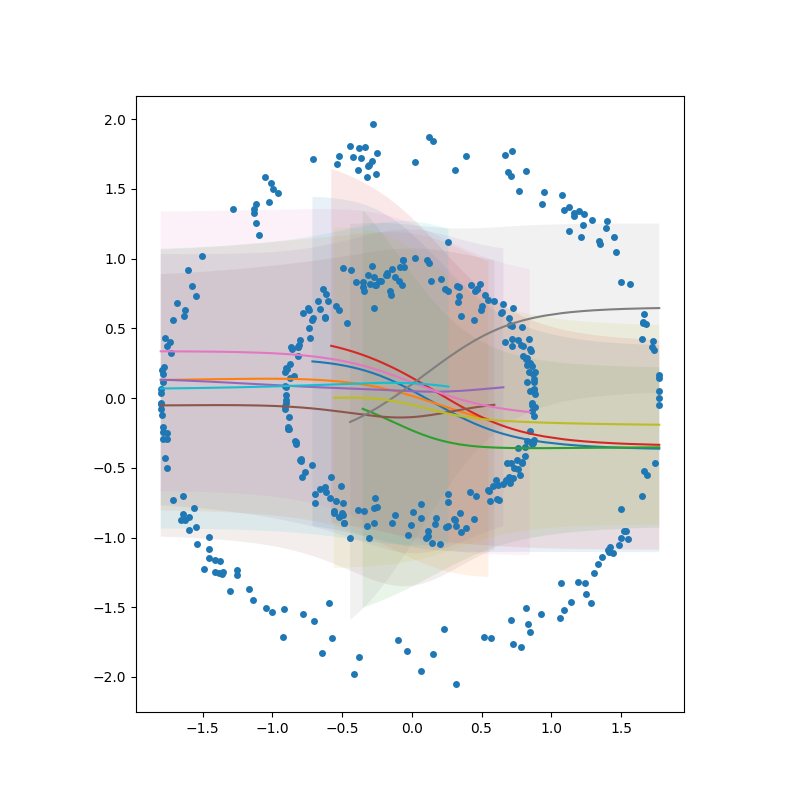}} &
		\makecell{\includegraphics[width=0.18\textwidth]{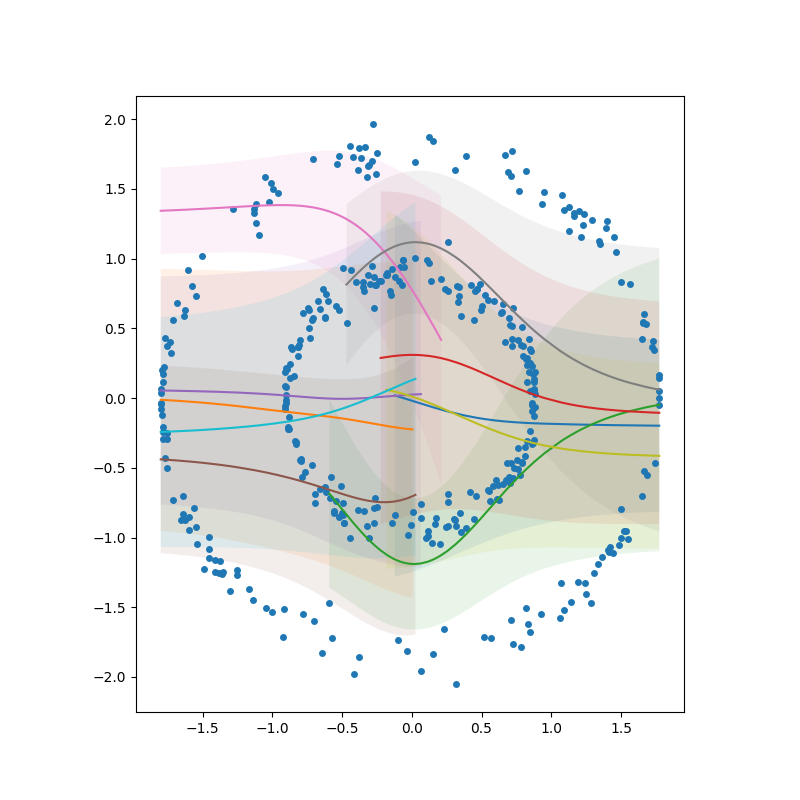}} &
	\makecell{\includegraphics[width=0.18\textwidth]{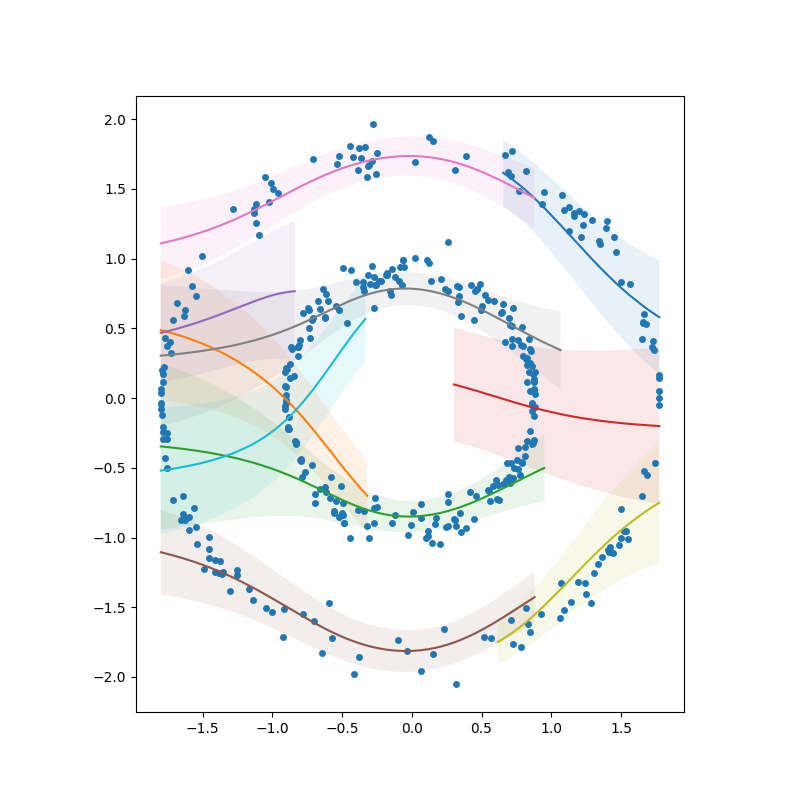}} &
	\makecell{\includegraphics[width=0.18\textwidth]{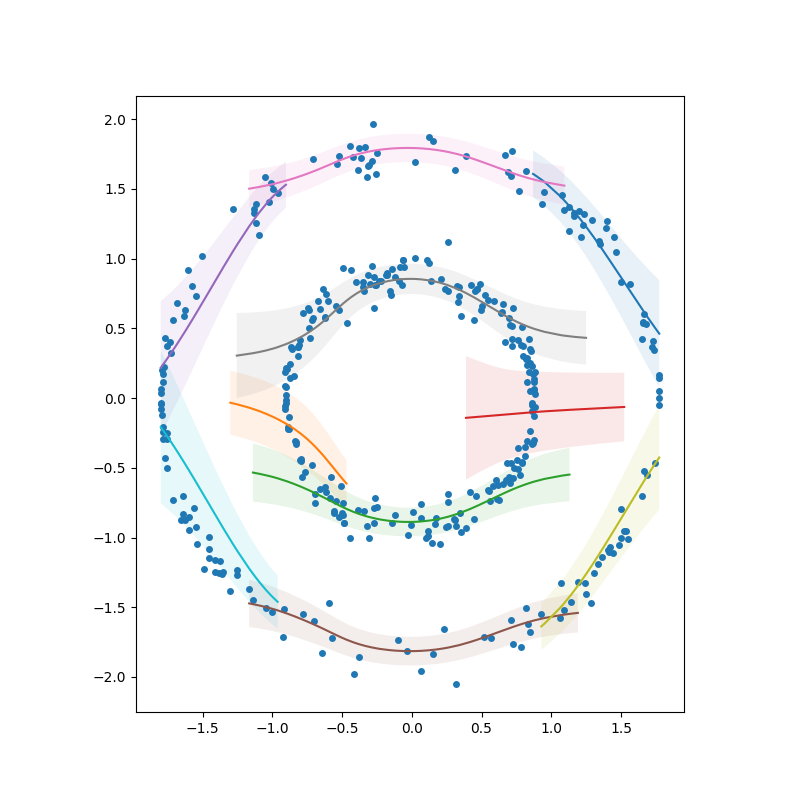}} &
	\makecell{\includegraphics[width=0.18\textwidth]{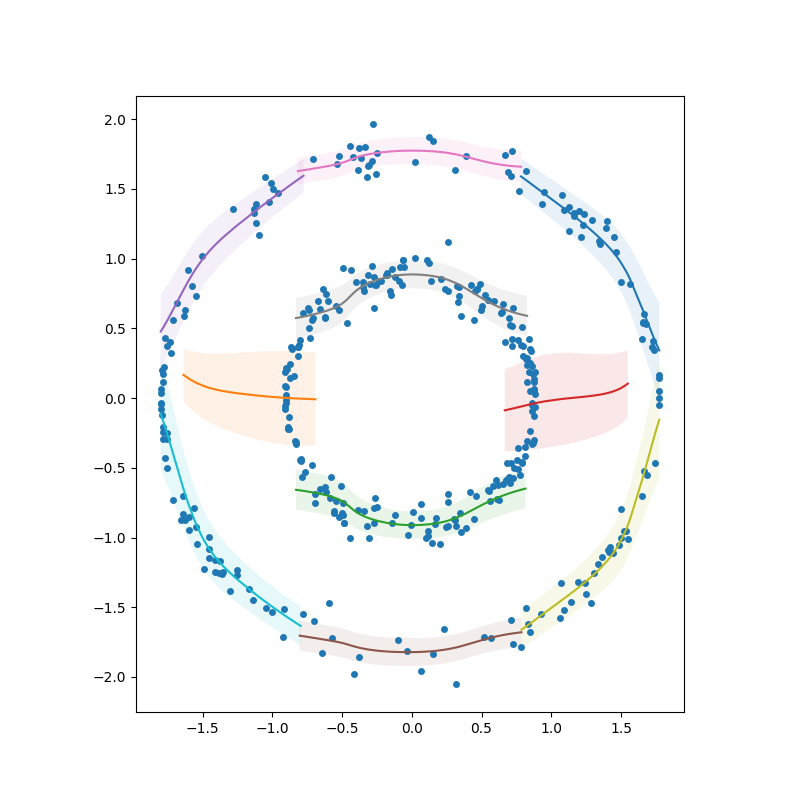}}\\
 		\end{tabular}
	\caption{Qualitative comparison between MDN and \our{} (our) on a simple synthetic dataset, as discussed in \cite{bishop1994mixture,pan2020implicit}. The objective was to utilize ten Gaussian components to cover a 2D shape comprising two concentric circles (indicated with blue dots). For each regression mode, the final values of the mean and standard deviation parameters are presented in the form of a range plot. The results presented are those obtained after 1, 4, 16, 64, 512, and 1024 epochs of training (from left to right). It can be observed that both methods demonstrate comparable performance, but \our{} exhibits a more rapid convergence in certain regions.}
\label{cluster_dis}
\end{figure*}

This paper presents CEC-MMR, a novel approach to multi-modal regression tasks that allows for automatic detection of the number of Gaussian components and, given an attribute and its value, enables the identification of that attribute with the underlying mode. Our solution is based on the Cross-Entropy Clustering (CEC) framework \cite{spurek2019online,spurek2017active,tabor2014cross}, rather than Gaussian Mixture Models (GMMs)~\cite{reynolds2009gaussian}. It is important to note that, in contrast to other approaches, CEC does not perform density estimation. In contrast, this method generates clusters defined by Gaussian distributions, with the upper bound on the number of clusters predetermined. The formation of clusters is an independent process, with each cluster having its own associated cost. Consequently, in the event that a component exhibits suboptimal quality, it will be eliminated (see Figure~\ref{img:cec_gmm}). 
It is of particular importance to emphasize that the distinction between \our{} and MDNs in favour of our algorithm is particularly evident when the upper bound for the number of Gaussian components is significantly larger than the actual number of nodes in a given dataset. While in some instances the results of the Mixture Density Network technique converge to the correct multi-modal regression, our algorithm does so in a more visually appealing manner. Consequently, we can conservatively bound the number of nodes with the assurance of obtaining high-quality results, in contrast to the case of the MDN loss function, where the aforementioned bound should be as precise as possible.

Our contribution can be summarized as follows:
\begin{itemize}
\item we present \our{}, a novel approach to multi-modal regression problems based on a learning procedure using a Cross-Entropy Clustering (CEC) objective function,
\item in contrast to classical Mixture Density Networks (MDNs), our method demonstrates the ability to automatically identify the number of Gaussian components and given an attribute
 and its value, to uniquely identify it with the underlying mode,
\item we conduct experiments on a range of synthetic and real-world datasets, which illustrate the enhanced performance of our approach in comparison to existing state-of-the-art methods.

\end{itemize}

\section{Related Work}
\label{Related-Works}

A common approach to multi-modal regression is to combine the outputs of a neural network with those of parametric distributions. This is exemplified by Mixture Density Networks (MDNs), which are employed to predict the parameters of Gaussian Mixture Distributions (GMMs) \cite{bishop1994mixture}. In this context, the output value is represented as a sum of numerous Gaussian random values, each with a distinct mean and standard deviation.
Alternative approaches (e.g., ~\cite{ambrogioni2017kernel,rothfuss2019noise,rothfuss2019conditional}), employ a Kernel Mixture Network (KMN) that integrates both non-parametric and parametric elements.  On the other hand, in \cite{guzman2012multiple} the authors introduce a Winner-Takes-All (WTA) loss for Support Vector Machines (SVMs) with multiple hypotheses as an output. This loss was applied to CNNs \cite{lee2016stochastic} for image classification, semantic segmentation, and image captioning. In turn, the authors of  \cite{pan2020implicit} proposed a multi-modal regression algorithm by employing the implicit function theorem to develop an objective for learning a joint parameterized function over inputs and targets.

On the other hand, a considerable number of approaches concentrate on the modeling of conditional probability. In particular, in \cite{lee2017desire} a framework for distant future prediction of multiple agents in complex scenes is presented. This method employs a conditional variational autoencoder (cVAE) to predict multiple long-term futures of interacting agents. In \cite{li2018flow}, the authors put forth a novel approach to motion encoding that incorporates a 3D cVAE. Similarly, in \cite{bhattacharyya2018bayesian} a novel approach to integrating dropout-based Bayesian inference into the cVAE is proposed. In turn, the authors of \cite{trippe2018conditional} present an efficient method for utilizing normalizing flows \cite{chen2018neural} as a flexible likelihood model for conditional density estimation. Specifically, they introduce a Bayesian framework for placing priors over conditional density estimators defined using normalizing flows and performing inference with variational Bayesian neural networks.

\section{Our Method}
\label{Our-Contribution}

In this section, we introduce \our{}, a novel approach to multi-modal regression tasks. We begin with a concise overview of classical Mixture Density Networks (MDNs), a prevalent tool for addressing such problems, and then proceed to elucidate the specifics of our proposed solution.

\begin{figure*}[ht!]
\centering
    \begin{tabular}{@{}c@{}c@{}c@{}c@{}c}
    \rotatebox[origin=c]{90}{\tiny MDN} &
\makecell{\includegraphics[width=0.3\textwidth]{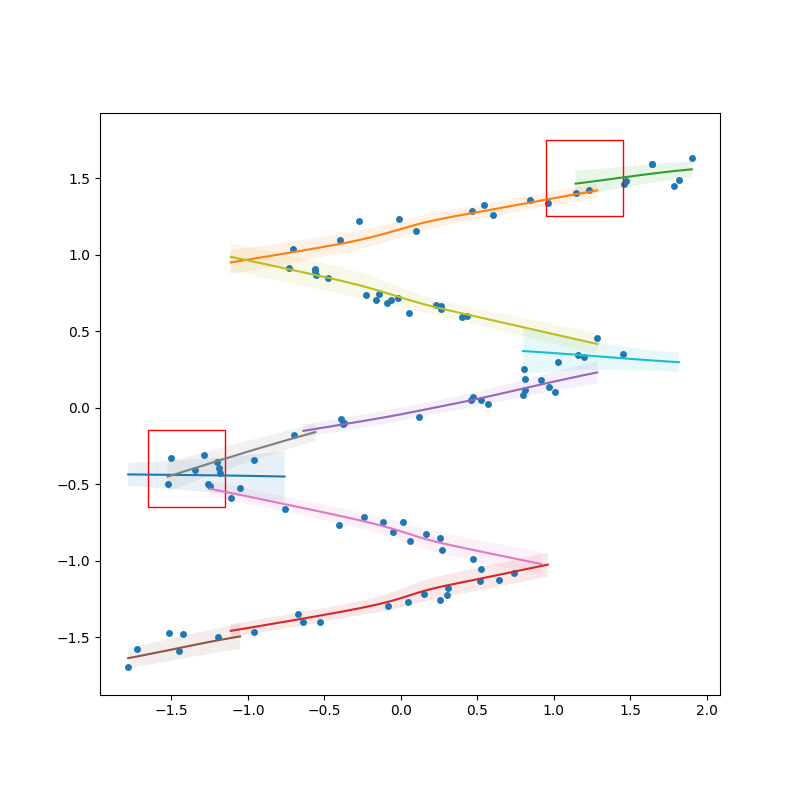}} & 
\makecell{\includegraphics[width=0.15\textwidth]{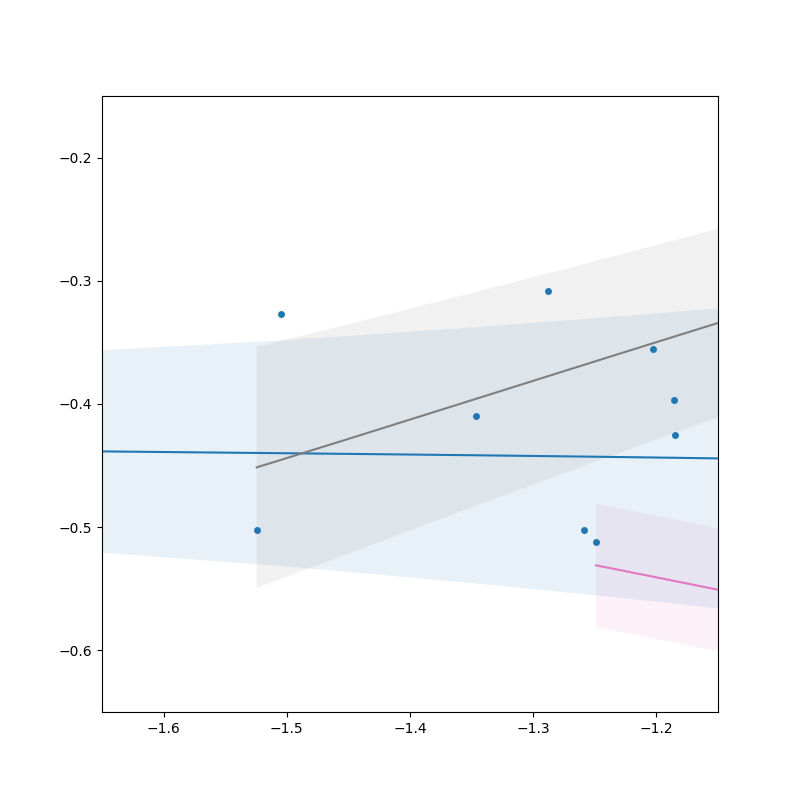}\\ \includegraphics[width=0.15\textwidth]{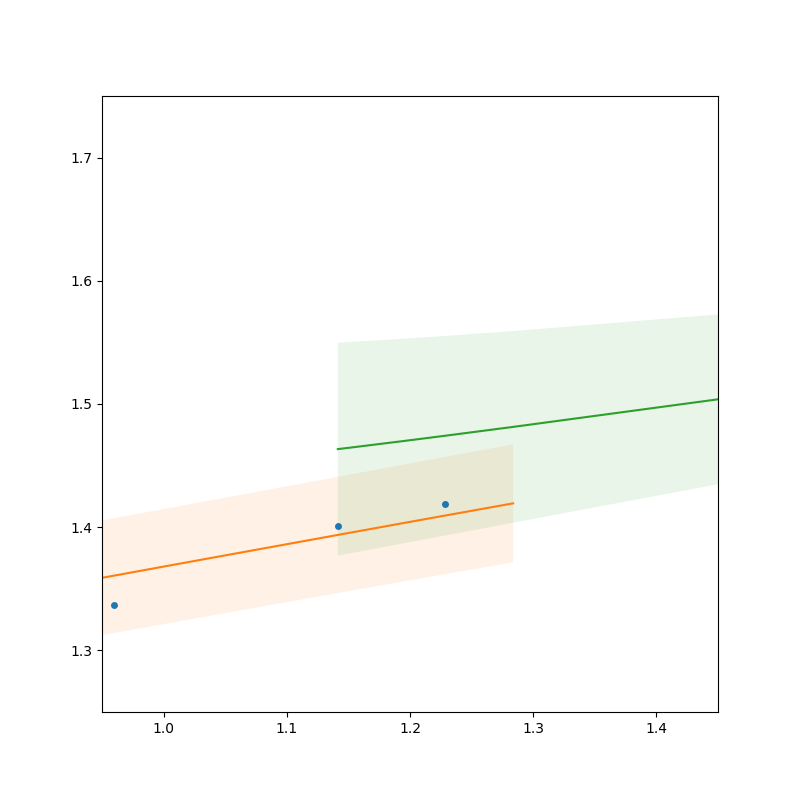}} &
    \makecell{\includegraphics[width=0.3\textwidth]{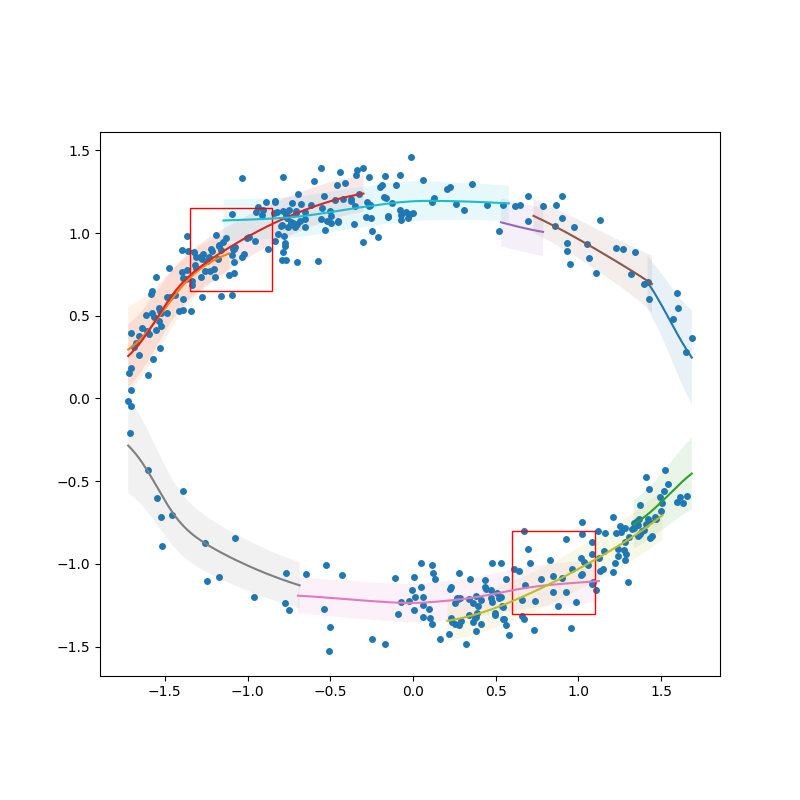}} & \makecell{\includegraphics[width=0.15\textwidth]{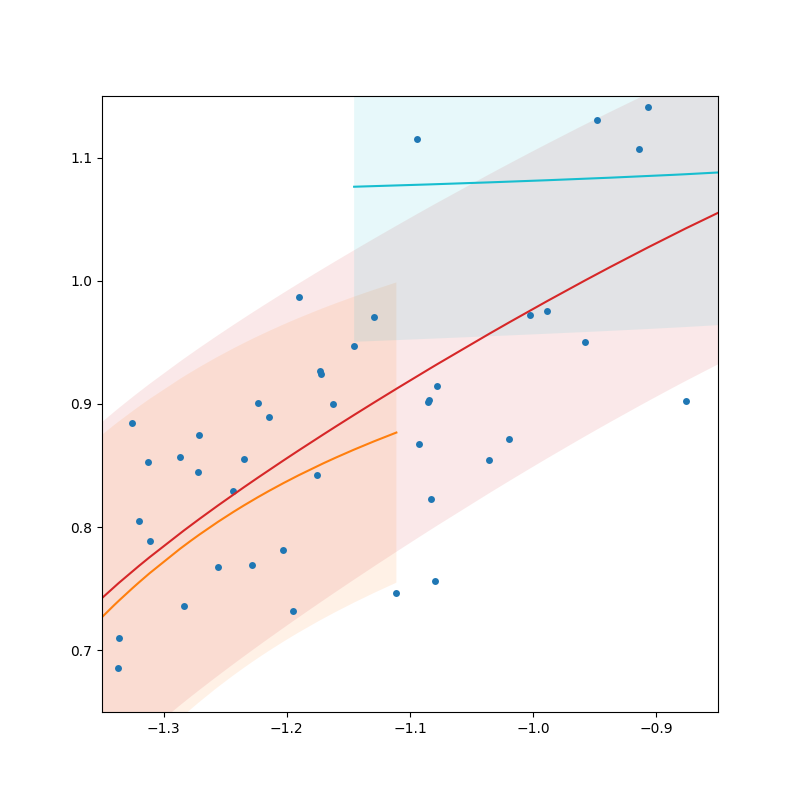}\\
    \includegraphics[width=0.15\textwidth]{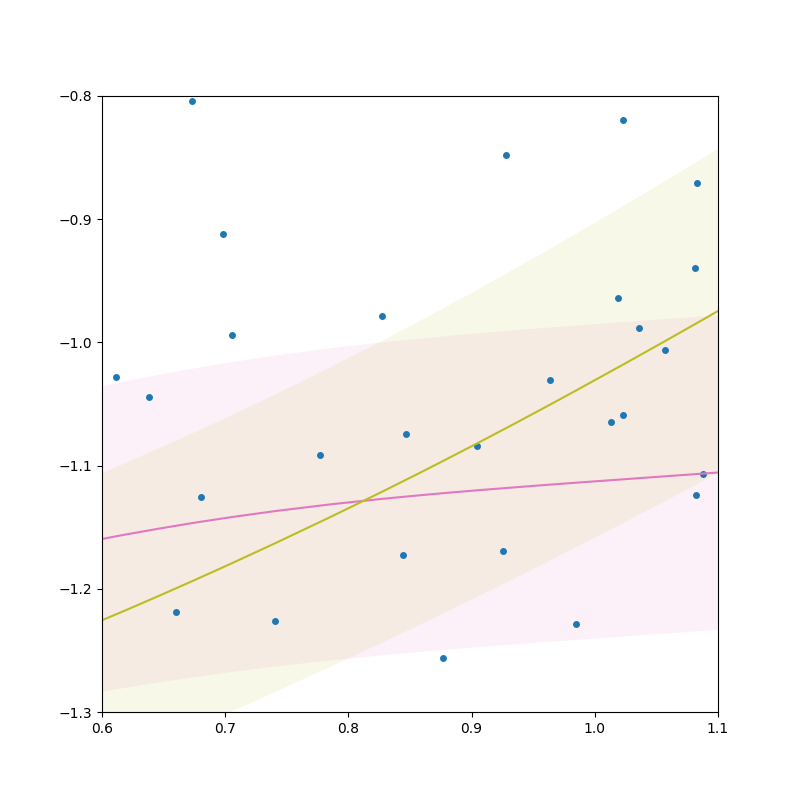} }\\
    \rotatebox[origin=c]{90}{\tiny \our{} (our)} &
    \makecell{\includegraphics[width=0.3\textwidth]{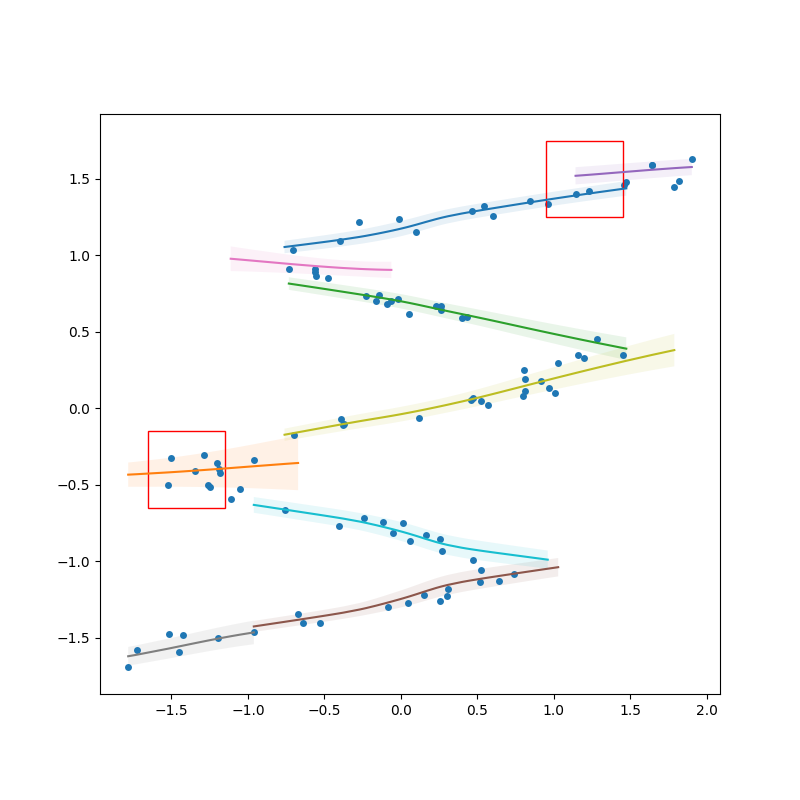}} & 
    \makecell{\includegraphics[width=0.15\textwidth]{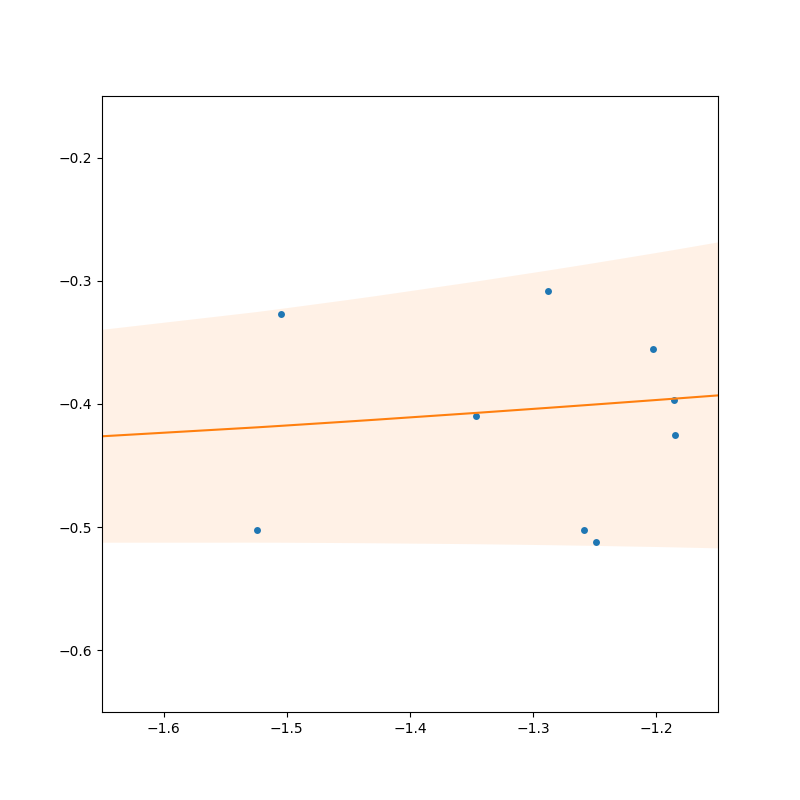}\\
    \includegraphics[width=0.15\textwidth]{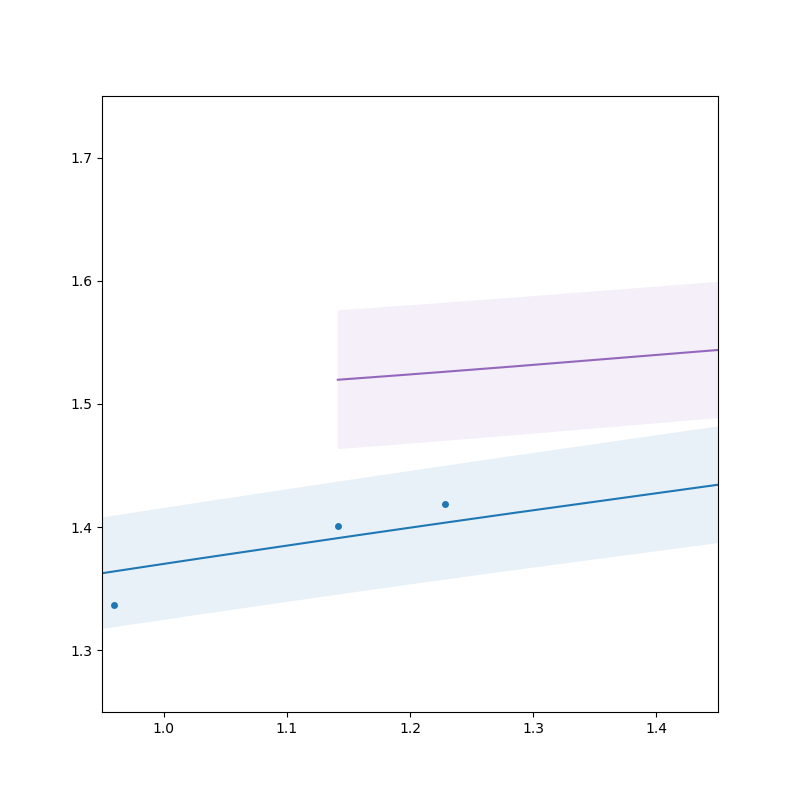}} &
    \makecell{\includegraphics[width=0.3\textwidth]{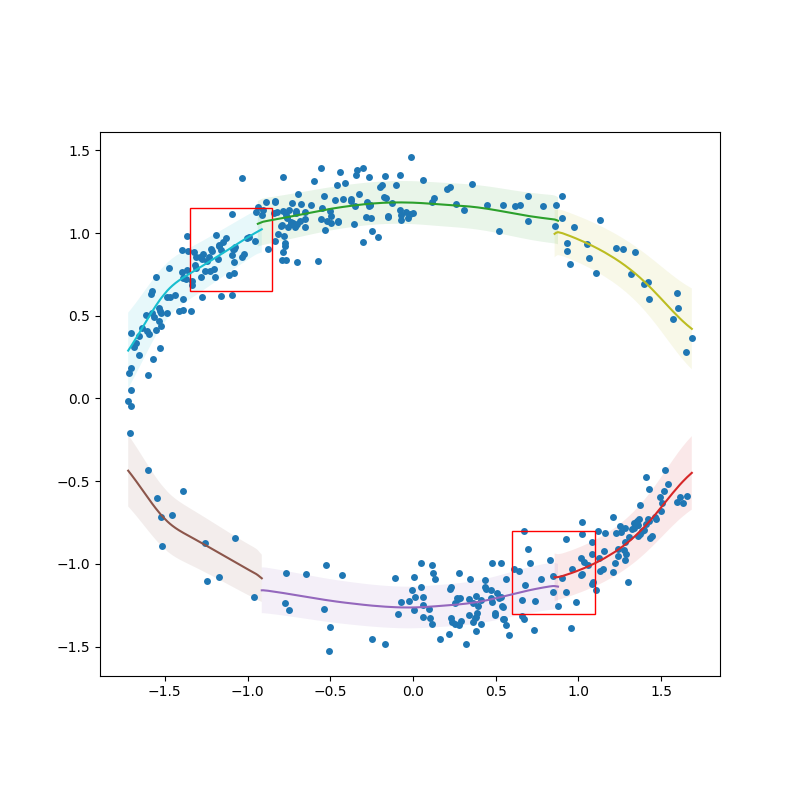}} &
    \makecell{\includegraphics[width=0.15\textwidth]{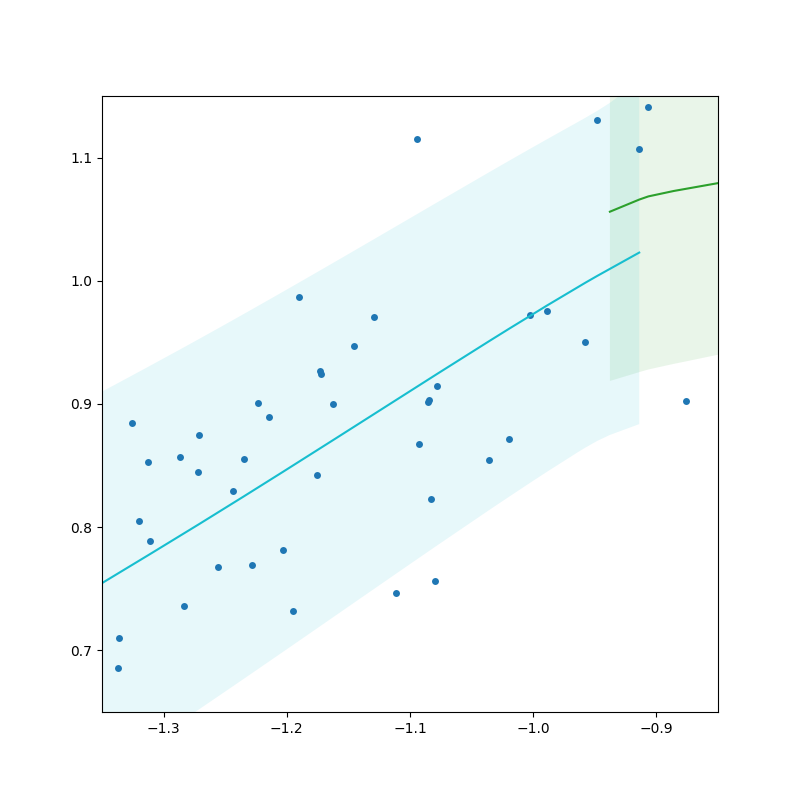}\\
    \includegraphics[width=0.15\textwidth]{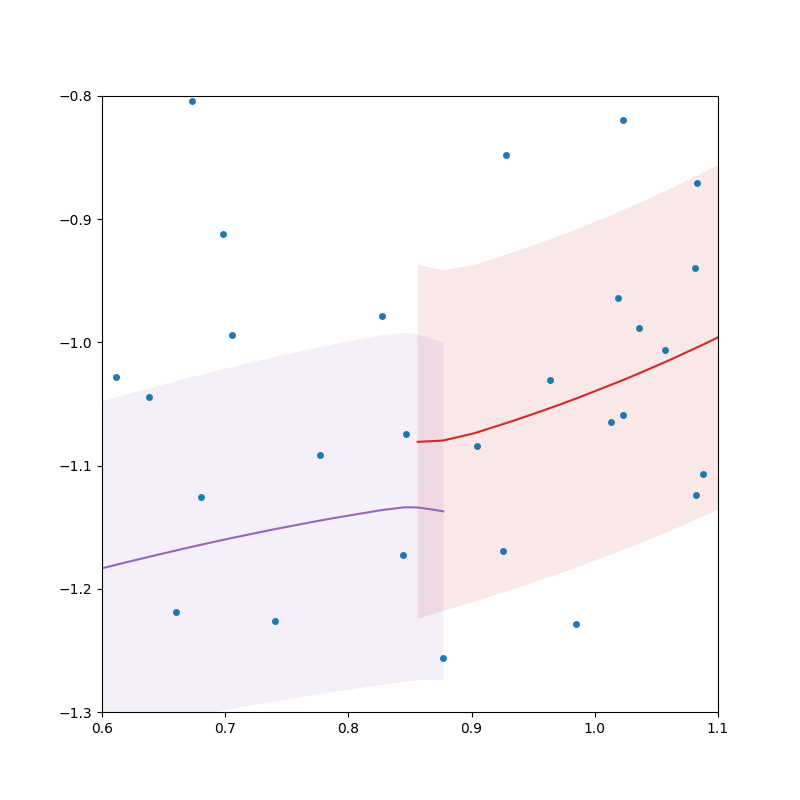}}
    \end{tabular}
\caption{Qualitative comparison between MDN and \our{} (our) on a simple synthetic dataset, as discussed in \cite{bishop1994mixture,pan2020implicit}.  The objective was to cover two 2D geometric shapes (indicated with blue dots), namely a zigzag (on the left) and an ellipse (on the right), using ten Gaussian components. For each regression mode, the final values of the mean and standard deviation parameters are presented in the form of a range plot. It can be observed that \our{} achieves superior accuracy compared to MDN, which is particularly evident in the regions indicated by red rectangles (see their zoomed versions on the right). Furthermore, our method was capable of reducing the number of Gaussians to 9 (for the zigzag shape data) and 6 (for the ellipse shape data).}
\label{img:2din1}
\end{figure*}

\subsection{Mixture Density Networks}
\label{MDN}

A typical approach to multi-modal regression problems, known as Mixture Density Networks (MDNs) \cite{bishop1994mixture}, is based on the use of mixture models with parameters learned by neural networks to approximate the conditional distribution of the output variable. In the majority of cases, MDNs utilize Gaussian Mixture Models (GMMs) \cite{reynolds2009gaussian} with probability density functions defined by the following formula:
\begin{equation}\label{eq:gmm}
\begin{array}{c@{\;}c}
p_{\text{GMM}}(y|x) = & \sum_{i=1}^k p_i(x) \mathcal{N}(\mu_i(x),\sigma_i(x))(y), 
\end{array}
\end{equation}
where $\mathcal{N}(\mu_i(x),\sigma_i(x))$ denotes the Gaussian distribution with the mean $\mu_i(x)$ and standard deviation $\sigma_i(x)$, and $p_i(x)$ is a positive constant (we assume that \mbox{$\sum_{i=1}^k p_i(x) = 1$}). Consequently, the conditional distribution $p(y|x)$ is estimated  during the optimization process, which involves maximizing of the following objective function:
\begin{equation}\label{eq:cost_gmm}
\begin{array}{c@{\;}c}
\mathcal{I}_{\text{GMM}}(Y|X) & =
\sum_{j=1}^n \sum_{i=1}^k p_i(x_j)  \mathcal{N}(\mu_i(x_j),\sigma_i(x_j))(y_j),
\end{array}
\end{equation}
where $Y|X=(y_j|x_j)_{j=1}^n$ represents the given samples of regression data, and $p_i$, $\mu_i$, and $\sigma_i$ are neural networks designed to model GMM's parameters. In this context, we utilize softmax activation for $p_i$, linear activation for $\mu_i$, and sigmoid activation for $\sigma_i$.

It is important to note that a significant drawback of MDNs is their inability to automatically adjust the number of components during the learning process. One potential solution is to employ the Cross-Entropy Clustering (CEC) framework \cite{tabor2014cross}, which results in the constitution of our proposed \our{} method, outlined in the following subsection.

\begin{figure*}[ht!]
	\centering
\begin{tabular}{c@{\;\;\;}c@{\;\;\;}c@{\;\;\;}c}
\makecell{\includegraphics[width=0.23\textwidth]{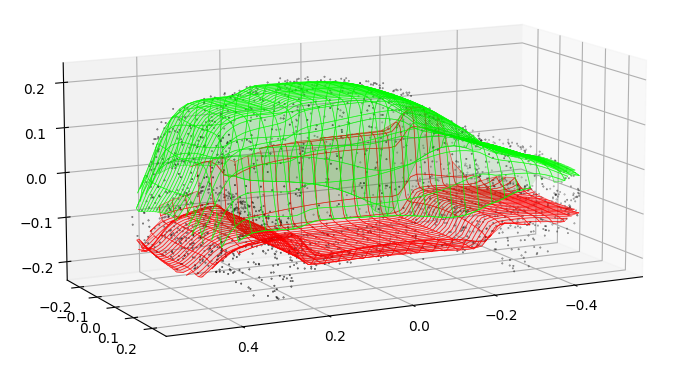}} &
 
\makecell{\includegraphics[width=0.23\textwidth]{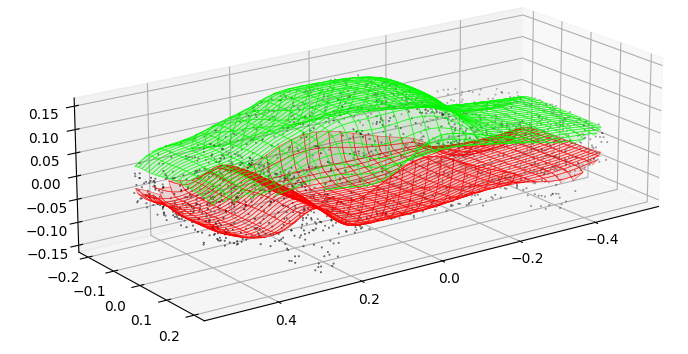}} &
 \makecell{\includegraphics[width=0.23\textwidth]{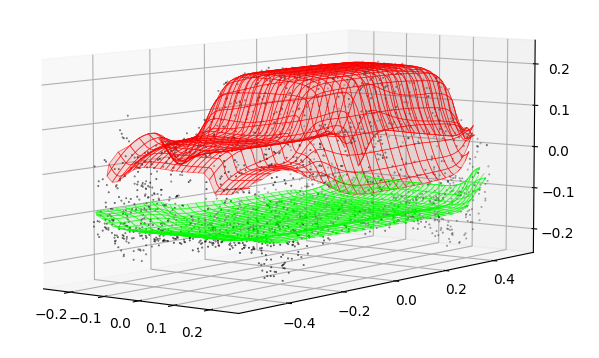}} &
\makecell{\includegraphics[width=0.23\textwidth]{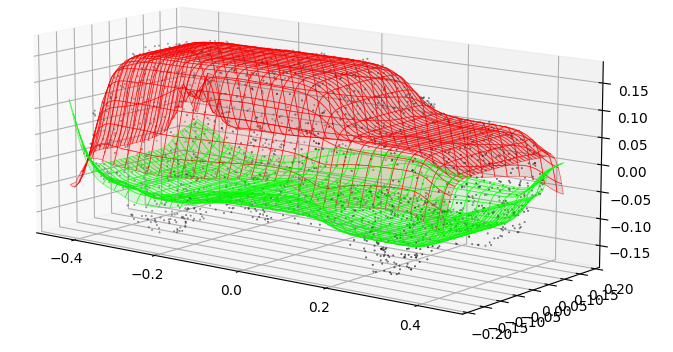}}
    \end{tabular}	
	\caption{Qualitative results of bimodal \our{} (our) for the approximation of four 3D car shapes. The examples presented (indicated as blue dots) were generated by sampling 2048 points from the meshes of selected objects from the ShapeNet dataset~\cite{chang2015shapenet}. Each 3D object was treated as a function from $\R^2$ to $\R$. It should be noted that our method is capable of successfully modeling two complementary components, namely the car chassis and the car body.
}
\label{img:3din}
\end{figure*}

\begin{algorithm*}[]
\label{alg}
\SetProcNameSty{textsc}
\SetProcArgSty{textsc}
\caption{ \our{}}
\SetKw{End}{end \\}
\SetKwRepeat{Do}{do}{while}
\SetKwInOut{Input}{Input}
\SetKwInOut{Output}{Output}
\Input{Initial network
\newline
$X$ -- input dataset
\newline
$Y$ -- output dataset
}
\Output{Resulting network, cl(x, k): $X \times \left \lbrace 1, \ldots, k \right \rbrace \mapsto \left \lbrace 0, 1 \right \rbrace$
}
\Begin{
    Split $(X,Y)$ into $(X_\text{train},Y_\text{train})$ and $(X_\text{test},Y_\text{test})$ \
    
    \For {$\text{epoch} \in \{ 1, \ldots, \text{no. of epochs} \}$ } {
        Split $(X_\text{train},Y_\text{train})$ into batches $(X_\text{train}^1,Y_\text{train}^1), \ldots, (X_\text{train}^{\text{no. of batches}},Y_\text{train}^{\text{no. of batches}})$
        
        \For { $ \text{batch} \in \{ 1, \ldots, \text{no. of batches} \}$ } {
        Perform ADAM backward propagation to minimize the value of the loss function
        $
        -\frac{1}{|Y_{\text{epoch}}|} \mathcal{I}_\text{CEC}(Y_{\text{epoch}}|X_{\text{epoch}})
        $
        }
    }
    \For {$x \in X$}{
        \For {$cluster \in \left \lbrace 1, \ldots, k \right \rbrace$}{
            \If {$p_{cluster}(x) > \epsilon$}{
                $cl(x, cluster) := 1$
            } \Else {
                $cl(x, cluster) := 0$ \\
                $p_{cluster}(x) := 0$
            }
        }
        Renormalize each number $p_i(x)$ for $i \in \left \lbrace 1, \ldots, k \right \rbrace$ so that they sum to $1$
    }
}
\end{algorithm*}

\subsection{\our{}}
\label{CEC-regression}
The fundamental concept underlying Cross-Entropy Clustering (CEC) \cite{tabor2014cross} is to utilize the maximum value of Gaussian components rather than their sum, as is the case in GMMs. This yields the following conditional function\footnote{It should be noted that, in contrast to GMMs, this is not a probability density function.}:
\begin{equation}\label{eq:ces}
\begin{array}{c@{\;}c}
p_{\text{CEC}}(y|x) = & \max\{ p_i(x) \mathcal{N}(\mu_i(x),\sigma_i(x))(y)\mid i=1,\ldots, k\},
\end{array}
\end{equation}
where $p_i(x)$ is a nonnegative number (we assume that $ \sum_{i=1}^k p_i(x) = 1$)
while the remaining parameters are the same as in the case of GMMs. Consequently, in our proposed \our{} method, we substitute the objective function presented in Eq.~\eqref{eq:cost_gmm} with the following formula:
\begin{equation}\label{eq:cec_cost}
\begin{array}{l}
\!\!\!\!\mathcal{I}_{\text{CEC}}(Y|X) =  
\\
\sum_{j=1}^n \max\{ p_i(x_j) \mathcal{N}(\mu_i(x_j),\sigma_i(x_j))(y_j)\mid i=1,\ldots, k\},
\end{array}
\end{equation}
where all the notation is derived from that used in Eq.~\eqref{eq:cost_gmm}.
The consecutive steps of the training procedure are shown in Algorithm~\ref{alg}. Note that after the training phase, we sweep over all points and clusters of the input dataset and check whether for a given point the probability of the cluster is greater than the predefined constant $\varepsilon>0$. If so, we mark such a cluster as inactive and set its probability to $0$. Finally, we renormalize all probabilities so that they all sum to $1$.

\begin{table*}[ht!]
\caption{Quantitative comparison of the performance of \our{} with that of other state-of-the-art methods on small datasets from the UCI repository \cite{lichman2013uci}. The experimental setup proposed in \cite{trippe2018conditional} is utilized, which incorporates a range of classical methods with varying parameter configurations. The results are presented in terms of the mean log-likelihood (higher is better) calculated across the entire dataset, averaged over 20 runs (standard deviations are also provided). For each run, 5 random train/test splits with a proportion of 80/20 were used, and the same weights for initialization were employed. All numbers were multiplied by $10^2$, and the cells with the three highest scores were colored red, orange, and yellow, respectively. It should be noted that when applied to 20 Gaussian components, our method achieves one of the top three scores in each case, clearly outperforming the classical MDN from which it was derived.}
\centering
\begin{tabular}{c c c c c c }
\toprule
\multirow{2}{*}{Method} & \multicolumn{5}{c}{Dataset} \\
\cmidrule(rr){2-6}
 & Boston & Concrete & Energy & Wine & Yacht \\
\midrule
MDN-2 & $-2.65 \pm 0.03$ & $-3.23 \pm 0.03$ & $-1.60 \pm 0.04$ & $-0.91 \pm 0.04$ & $-2.70 \pm 0.05$ \\
MDN-5 & $-2.73 \pm 0.04$ & $-3.28 \pm 0.03$ & $-1.63 \pm 0.06$ & $ 1.43 \pm 0.07$ & $-2.54 \pm 0.10$ \\
MDN-20 & $-2.74 \pm 0.03$ & $-3.27 \pm 0.02$ & $-1.48 \pm 0.04$ & $ 1.21 \pm 0.06$ & $-2.76 \pm 0.07$ \\
LV-5 & $-2.56 \pm 0.05$ & $-3.08 \pm 0.02$ & $-0.79 \pm 0.02$ & $-0.96 \pm 0.01$  & $-1.15 \pm 0.05$ \\
LV-15 & $-2.64 \pm 0.05$ & $-3.06 \pm 0.03$ & $-0.74 \pm 0.03$ & $-0.98 \pm 0.02$ & $-1.01 \pm 0.04$ \\
NF-2 & $-2.40 \pm 0.06$ & $-3.03 \pm 0.05$ & \redc $-0.44 \pm 0.04$ & $-0.87 \pm 0.02$ & \orangec $-0.30 \pm 0.04$ \\
NF-5 & \yellowc $-2.37 \pm 0.04$ & $-2.97 \pm 0.03$ & $-0.67 \pm 0.15$ & $-0.76 \pm 0.10$ & $\redc -0.21 \pm 0.09$ \\
HMC & \redc $-2.27 \pm 0.03$ & \redc $-2.72 \pm 0.02$ & $-0.93 \pm 0.01$ & $-0.91 \pm 0.02$ & $-1.62 \pm 0.02$ \\
Dropout & $-2.46 \pm 0.25$ & $-3.04 \pm 0.09$ & $-1.99 \pm 0.09$ & $-0.93 \pm 0.06$ & $-1.55 \pm 0.12$ \\
MF & $-2.62 \pm 0.06$ & $-3.00 \pm 0.03$ & \orangec $-0.57 \pm 0.04$ & $-0.97 \pm 0.01$ & $-1.00 \pm 0.10$ \\
\our{}-2 & $-2.42 \pm 0.09$ & \yellowc $-2.86 \pm 0.04$ & $-0.94 \pm 0.03$ & \yellowc $2.48 \pm 0.03$ & $-1.03 \pm 0.02$ \\
\our{}-5 & $-2.38 \pm 0.05$ & \yellowc $-2.86 \pm 0.04$ & $-0.72 \pm 0.05$ & \orangec $7.52 \pm 0.37$ & $-0.84 \pm 0.07$ \\
\our{}-20 & \orangec $-2.33 \pm 0.04$ & \orangec $-2.8 \pm 0.04$ & \yellowc $-0.58 \pm 0.05$ & \redc $7.8 \pm 0.05$ & \yellowc $-0.66 \pm 0.12$ \\
\bottomrule
\end{tabular}
\label{tab:log}
\vspace{2cm}
\end{table*}

The utilization of the cost function presented on Eq.~\eqref{eq:cec_cost} has profound and far-reaching implications. As illustrated in \cite{tabor2014cross}, the incorporation of this modified objective introduces supplementary costs to each mode, thereby encouraging models with a minimal number of modes. Moreover, it permits the automatic reduction of the number of modes during the training procedure (via dropping components with $p_i$ below the established threshold), which, for MDNs, could only be accomplished through a manual process. In addition, by using the maximum instead of the sum, we can uniquely identify given data points with the underlying modes.

\section{Experiments}
\label{Experiments}

In this section, we present experimental evidence of the efficiency of our method in a variety of regression tasks, performed on both synthetic and real-world datasets. We start with a qualitative study on toy datasets consisting of 2D shapes (see \cite{bishop1994mixture,pan2020implicit}), and then proceed to a quantitative evaluation on six small UCI datasets, as proposed in \cite{trippe2018conditional}. Finally, we conduct experiments on the real-world Bike Sharing and Song Year datasets, inspired by those in \cite{pan2020implicit}, which rely on predicting the number of rental bikes in a given hour, given 114 preprocessed features, and the release year of a song, using respective audio features.

\subsection{Qualitative results on toy datasets}

We provide a qualitative comparison of \our{} with MDN in the approximation of simple 2D geometric shapes, as proposed in \cite{bishop1994mixture,pan2020implicit}. The results are presented in Figures \ref{cluster_dis} and \ref{img:2din1}. As can be observed, our model demonstrates superior performance in terms of accuracy and convergence.

Moreover, Figure~\ref{img:3din} demonstrates the implementation of bimodal CEC-MMR for the approximation of four 3D car shapes, represented as points uniformly sampled on the meshes of selected objects from the ShapeNet database \cite{chang2015shapenet}. Notably, our method effectively discriminates between the car chassis and the car body.

\subsection{Quantitative results on synthetic datasets}

We compare the performance of \our{} with that of other state-of-the-art approaches in solving multi-modal regression tasks on six small datasets from the UCI repository~\cite{lichman2013uci} (namely: Boston, Concrete, Energy, Power, Wine, and Yacht). We utilize the experimental setup proposed in \cite{trippe2018conditional}, which incorporates a range of classical methods with varying parameter configurations. In particular, for MDN and \our{} we considered settings with 2, 5, or 20 Gaussian components, for the Latent Variable (LV) input neural network the results were computed with 5 and 15 samples of noise, for the Normalizing Flow (NF) we applied 2 and 5 radial warpings, and for the Bayesian neural network model with homoscedastic Gaussian likelihood we used two approximate inference methods, namely the Mean Field (MF) variational approximation and the Hamiltonian Monte Carlo (HMC).


\begin{table*}[ht!]
\centering
\caption{Quantitative evaluation of \our{} (our) on the Bike Sharing dataset~\cite{fanaee2014event}, based on the experimental framework proposed in \cite{pan2020implicit}, which includes a number of state-of-the-art approaches. We present the obtained prediction accuracy in terms of Root Mean Square Error (RMSE, lower is better) and Mean Absolute Error (MAE, lower is better), calculated separately for the train and test subsets of the considered dataset and averaged over 5 runs (standard deviations are also provided). For each run, 20 random train/test splits were used with a ratio of 90/10, and the same weights for initialization were employed. All numbers were multiplied by $10^2$, and the cells with the three highest scores were colored red, orange, and yellow, respectively. It is noteworthy that \our{} performs better than the other methods on the train dataset.}
\begin{tabular}{ccccccccc}
\toprule
\multirow{2}{*}{Method} & \multicolumn{4}{c}{Dataset/Metric}\\
\cmidrule(rr){2-5}
 & Train/RMSE & Train/MAE & Test/RMSE & Test/MAE \\
\midrule
LinearReg & $10094.40 \pm 13.60$ & $7517.64 \pm 19.95$ & $10129.40 \pm59.26$ & $7504.22 \pm 44.20$ \\
LinearPoisson & $8798.26 \pm 14.58$ & $5920.99 \pm 13.66$ & $8864.90 \pm 66.07$ & $5935.00 \pm 38.32$ \\
NNPoisson & $\orangec 1620.46 \pm 47.71$ & $\orangec 1071.39 \pm 29.55$ & $4150.03 \pm 77.76$ & $2616.49 \pm 20.45$ \\
L2 & $1919.74 \pm 23.12$ & $1421.36 \pm 21.35$ & \yellowc $3726.40 \pm 49.51$ & $2526.77 \pm 27.89$ \\
Huber & \yellowc $1914.34 \pm 33.87$ & \yellowc $1398.41 \pm 21.11$ & $\orangec 3675.03 \pm 40.67$ & $2487.61 \pm 11.05$ \\
MDN & $2888.06 \pm 64.55$ & $1456.31 \pm 76.21$ & $3948.48 \pm 63.95$ & $\redc 2298.47 \pm 36.37$ \\
MDN(worst) & $3506.32 \pm 166.30$ & $1604.43 \pm 35.90$ & $4452.52 \pm 166.65$ & \yellowc $2431.40 \pm 41.31$ \\
Implicit & $2075.33 \pm 7.01$ & $1504.63 \pm 4.34$ & \redc $3674.08 \pm 16.82$ & $\orangec 2419.54 \pm 12.22$ \\
Implicit(worst) & $2154.70 \pm 7.55$ & $1602.03 \pm 5.35$ & $3749.95 \pm 16.36$ & $2514.99 \pm 13.51$ \\
\our{} (our) & $\redc 1378.77 \pm 139.09$ & $\redc 796.01 \pm 43.21$ & $4104.0 \pm 557.87$ & $2669.06 \pm 258.15$ \\
\bottomrule
\end{tabular}
%
\label{tab:bike}
\end{table*}

Table~\ref{tab:log} presents the results of the conducted experiments in terms of the mean log-likelihood calculated across the entire dataset\footnote{It is important to note that in the case of \our{}, the likelihood is to be computed as the weighted sum of Gaussian likelihoods, rather than as a maximum. However, it is likely that the number of components has been reduced during the training process.}. It should be noted that our method, when applied with 20 Gaussian components, achieves one of the highest three scores in each case, clearly outperforming the classical MDN from which it was derived.

\subsection{Quantitative results on real-world datasets}

We assess the effectiveness of our method in addressing the problem of multi-modal regression for real-world data from the Bike Sharing dataset~\cite{fanaee2014event} and the Song Year dataset~\cite{bertin2011million}. The evaluation is based on the experimental framework proposed in \cite{pan2020implicit}, encompassing a range of state-of-the-art approaches.

Tables \ref{tab:bike} and \ref{tab:song} demonstrate the obtained prediction accuracy in terms of Root Mean Square Error (RMSE) and Mean Absolute Error (MAE), calculated separately for the train and test subsets of both considered datasets. It is notable that in each case, \our{} exhibits superior performance with respect to the other methods on the train dataset. Additionally, it attains the highest metric scores when applied to the test data from the Song Year dataset.


\begin{table*}[]
\centering
\caption{Quantitative evaluation of \our{} (our) on the Song Year dataset~\cite{bertin2011million}, based on the experimental framework proposed in \cite{pan2020implicit}, which includes a number of state-of-the-art approaches. We present the obtained prediction accuracy in terms of Root Mean Square Error (RMSE, lower is better) and Mean Absolute Error (MAE, lower is better), calculated separately for the train and test subsets of the considered dataset and averaged over 5 runs (standard deviations are also provided). For each run, 5 random train/test splits were used with a ratio of 80/20, and the same weights for initialization were employed. All numbers were multiplied by $10^2$, and the cells with the three highest scores were colored red, orange, and yellow, respectively. It is noteworthy that our method achieves the superior metric scores on both the train and test datasets.}
\begin{tabular}{ccccccccc}
\toprule
\multirow{2}{*}{Method} & \multicolumn{4}{c}{Dataset/Metric}\\
\cmidrule(rr){2-5}
 & Train/RMSE & Train/MAE & Test/RMSE & Test/MAE \\
\midrule
LinearReg & $956.40 \pm 0.37$ & $681.56 \pm 0.68$ & $957.56 \pm1.49$ & $681.66 \pm 1.52$ \\
L2 & \orangec $850.20 \pm 2.50$ & \yellowc $590.18 \pm 1.63$ & \yellowc $895.77 \pm 2.75$ & $608.42 \pm 1.03$ \\
Huber & \yellowc $872.56 \pm 5.00$ & \orangec $569.49 \pm 1.75$ & $898.33 \pm 2.67$ & \orangec $581.48 \pm 1.37$ \\
MDN & $955.85 \pm 12.97$ & $605.58 \pm 4.02$ & $957.76 \pm 13.53$ & $615.57 \pm 4.37$ \\
MDN(worst) & $1699.48 \pm 239.09$ & $1212.82 \pm 230.93$ & $1700.11 \pm 240.09$ & $1215.60 \pm 231.33$ \\
Implicit & $876.71 \pm 1.88$ & $598.68 \pm 4.11$ & \orangec $890.61 \pm 2.87$ & \yellowc $606.91 \pm 3.48$ \\
Implicit(worst) & $886.83 \pm 2.92$ & $604.16 \pm 2.08$ & $896.08 \pm 3.00$ & $612.25 \pm 2.53$ \\
\our{} (our) & \redc $542.87 \pm 11.62$ & \redc $370.04 \pm 19.9$ & \redc $578.84 \pm 14.18$ & \redc $373.94 \pm 18.85$ \\
\bottomrule
\end{tabular}
%
\label{tab:song}
\end{table*}

\subsection{Implementation details}
We implemented our algorithm using the fully connected neural network with the same number of neurons in each hidden layer, the tangent hyperbolic activation function, the batch normalization after each layer, and the dropout rate of $0.2$. The values of the following parameters: number of hidden layers, number of neurons in each hidden layer, batch size, learning rate, and type of loss function were determined individually for each dataset and experiment. For the simple datasets from the UCI repository, we used the following ranges of parameters: $\{1,2,3\}$ for the number of hidden layers, $\{16,32,64\}$ for the number of neurons in each hidden layer, $\{16,32,64\}$ for the batch size, $\{0.01, 0.001, 0.0001, 0.00001\}$ for the learning rate. On the other hand, for the Bike Sharing and Song Year datasets, we performed the separate grid search with the following ranges of parameters: $\{1,2,3\}$ for the number of hidden layers, $\{32,64,128\}$ for the number of neurons in each hidden layer, $\{64,128,256\}$ for the batch size, $\{0.01, 0.001, 0.0001, 0.00001\}$ for the learning rate. The network parameters in all experiments and datasets were optimized using the Adam optimizer with the following hyperparameters: $\beta_1 = 0.9$ and $\beta_2 = 0.99$.

\section{Conclusions}
\label{Conclusions}

In the paper, we introduced \our{}, which represents a novel approach to multi-modal regression problems. Our method is based on a learning procedure that employs a Cross-Entropy Clustering (CEC) objective function instead of a mixture of Gaussians, which is utilized by classical Mixture Density Networks (MDNs).  Consequently, \our{} enables the automatic identification of the number of Gaussian components and the efficient discrimination between them when attempting to capture a specific subset of data. The results of the experiments that were conducted demonstrate the superiority of our approach to state-of-the-art methods for solving multi-modal regression tasks on both synthetic and real-world datasets.

\subsection{Limitations} The primary limitation of \our{} is determining an a priori value of the threshold used to reduce the redundant Gaussian components of our multi-modal regression model. Indeed, an alternative method has already been developed that, despite offering reduced flexibility, enables the explicit identification of relevant components without prior knowledge of hyper-parameters. However, the results observed in our experimental trials did not yet meet the desired standard, and thus, this approach is being pursued as a potential direction for further research.

\section*{Acknowledgements}

The work of M. Mazur was supported by the National Centre of Science (Poland) Grant No. 2021/41/B/ST6/01370. 
The work of P. Spurek was supported by the National Centre for Science (Poland), Grant No 2023/50/E/ST6/00068.
Some experiments were performed on servers purchased with funds from the flagship project entitled ``Artificial Intelligence Computing Center Core Facility'' from the DigiWorld Priority Research Area within the Excellence Initiative -- Research University program at Jagiellonian University in Kraków.







\bibliographystyle{splncs04}


\end{document}